\documentclass[letterpaper, 10 pt, conference]{ieeeconf}  % Comment this line out if you need a4paper

\IEEEoverridecommandlockouts                              % This command is only needed if 
\usepackage{graphicx} % for pdf, bitmapped graphics files
\usepackage{amsmath} % assumes amsmath package installed
\usepackage{amssymb}  % assumes amsmath package installed
\usepackage{cite}
\usepackage{textcomp}
\usepackage{tabularx}
\usepackage{gensymb}
\usepackage{breqn}
\usepackage{placeins}
\usepackage[]{multirow}
\usepackage{subcaption}
\title{\LARGE \bf
Simulation-based Testing for Early Safety-Validation of Robot Systems
}

\newcommand{\breakingcomma}{%
	\begingroup\lccode`~=`,
	\lowercase{\endgroup\expandafter\def\expandafter~\expandafter{~\penalty0 }}}

\author{Tom P. Huck, Christoph Ledermann, and Torsten Kr\"oger% <-this % stops a space
\thanks{\noindent This work was funded by the Ministry of Economics, Work and Housing of the State of Baden-Württemberg in the research project 'RoboShield'.}% <-this % stops a space
\thanks{The authors are with the Intelligent Process Automation and Robotics Lab, Institute of Anthropomatics and Robotics (IAR-IPR), Karlsruhe Institute of Technology (KIT), 76131 Karlsruhe, Germany. Corresponding author: Tom Huck
({\tt\small tom.huck@kit.edu})\newline © 2020 IEEE.  Personal use of this material is permitted.  Permission from IEEE must be obtained for all other uses, in any current or future media, including reprinting/republishing this material for advertising or promotional purposes, creating new collective works, for resale or redistribution to servers or lists, or reuse of any copyrighted component of this work in other works.}%
}

\begin{document}

\maketitle
\thispagestyle{empty}
\pagestyle{empty}

%%%%%%%%%%%%%%%%%%%%%%%%%%%%%%%%%%%%%%%%%%%%%%%%%%%%%%%%%%%%%%%%%%%%%%%%%%%%%%%%
\begin{abstract}
Industrial human-robot collaborative systems must be validated thoroughly with regard to safety. The sooner potential hazards for workers can be exposed, the less costly is the implementation of necessary changes. Due to the complexity of robot systems, safety flaws often stay hidden, especially at early design stages, when a physical implementation is not yet available for testing. Simulation-based testing is a possible way to identify hazards in an early stage. However, creating simulation conditions in which hazards become observable can be difficult. Brute-force or Monte-Carlo-approaches are often not viable for hazard identification, due to large search spaces. This work addresses this problem by using a human model and an optimization algorithm to generate high-risk human behavior in simulation, thereby exposing potential hazards. A proof of concept is shown in an application example where the method is used to find hazards in an industrial robot cell.
%Industrial robot systems must be validated thoroughly with regard to safety. The sooner potential hazards can be exposed, the less costly changes have to be made at a later development stage. Due to the complexity of robot systems safety flaws often stay hidden, especially at early design stages, when a physical implementation is not yet available for testing. Simulation-based testing is a possible way to identify hazards in an early stage. However, creating simulation conditions in which hazards become observable can be difficult, and Brute Force or Monte Carlo approaches are often not viable due to large search spaces. This work uses a human model and an optimization algorithm to generate high-risk human behavior in simulation, thereby exposing potential hazards. A proof of concept is shown in an application example where the method is used to find hazards in an industrial robot cell.
\end{abstract}

%%%%%%%%%%%%%%%%%%%%%%%%%%%%%%%%%%%%%%%%%%%%%%%%%%%%%%%%%%%%%%%%%%%%%%%%%%%%%%%%
\section{INTRODUCTION}
\label{sec:Introduction}
\begin{figure}[t]
	\includegraphics[width=\columnwidth]{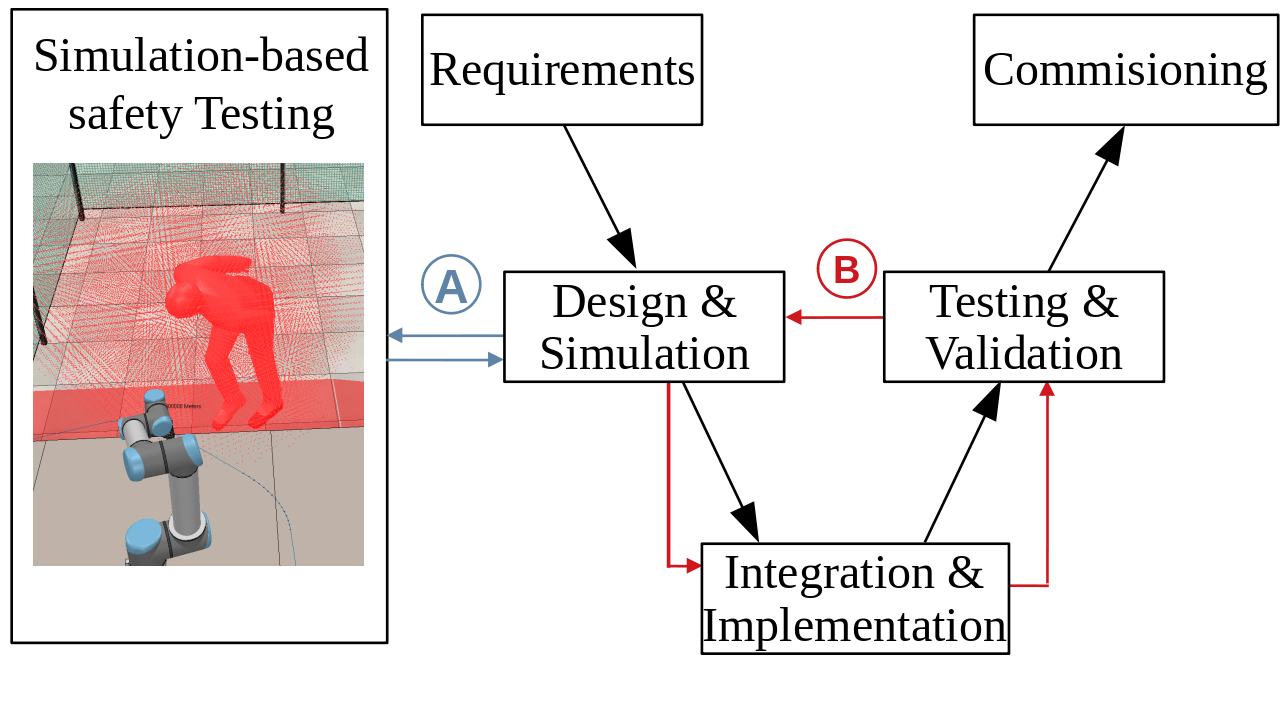}
	\caption{\small Simplified model of a development process to illustrate the benefits of simulation-based testing: Hazards identified early through simulation-based safety testing (A, blue) require smaller (and thus, less costly) iterations in the development process than hazards identified in the testing- and validation-phase (B, red). Although simulation-based testing cannot replace the testing- \& validation phase, it can reduce the need for costly iterations in the development process.}
	\label{fig:V-Model}
\end{figure}

Industrial Human-Robot Collaboration (HRC) promises a more flexible production and a more direct support for human workers \cite{Mueller2016}. In HRC applications, human and robot work in close vicinity or even in direct collaboration. Safety fences, which have traditionally been used to ensure safety of human workers, are (at least partially) absent. Instead, sensor- and software-based safety measures, such as laser scanners, light curtains, velocity limitation and collision detection, are used to ensure that the robot system does not pose any hazard to human workers. Safety flaws in the configuration of these safety measures can lead to hazards. Thus, a thorough safety-validation is required. Furthermore, ISO 10218-2, the safety standard for industrial robot systems, specifically states that prior to commissioning, a risk assessment must be conducted to identify and assess potential hazards \cite{ISO10218-2}.\newline
The sooner a hazard is uncovered in the development process, the less corrective changes to the system have to be made later. Since early changes require smaller iterations in the development process and thus are less costly (compare Fig. \ref{fig:V-Model}), it is desirable to identify hazards as early as possible. In early development stages there is usually no physical implementation available that could be used for this purpose. Instead, early development stages typically rely on simulation models, e.g. for planning the cell layout or optimizing the workflow. It would be beneficial to use these simulation models also for the early identification of potential hazards. However, to find hazards in simulation, one must overcome a major challenge: In many cases, hazards are \textit{hidden}. This means that there are certain safety-critical flaws in the design of the system which may result in hazards, but only become manifest in specific situations. In a dynamic simulation, it can be very difficult to find simulation sequences that uncover these hazardous situations, especially when the simulation is highly detailed.\newline
A recent promising approach to this problem is the concept of Adaptive Stress Testing (AST) \cite{Lee2018}. AST exposes hazards with a reinforcement learning agent that creates adversarial testing conditions. AST was successfully applied in several safety-critical domains such as aerospace engineering \cite{Lee2015} and autonomous driving \cite{Koren2018}. In this paper, we show how AST can be applied to find hazards in robot systems. We use a Monte-Carlo Tree Search (MCTS) algorithm to control a virtual human model which we place in a simulation model of the robot system. The MCTS acts as an optimization algorithm that adapts human behavior to maximize a risk metric, thereby creating high-risk situations which are more likely to uncover hazards. In other words, the human model exposes hazards by \textit{learning to provoke hazardous situations} in simulation. Although this approach cannot guarantee to find all existing hazards, it can help to uncover hazards which would have been overlooked otherwise, especially those that only become apparent in very specific situations.% A hazard identification tool that is based on this approach would therefore be a valuable addition to existing hazard identification methods. 
%The remainder of this paper is structured as follows: Section \ref{sec:RelatedWork} gives an overview of HRI-related risk assessment and hazard identification methods. Section \ref{sec:Approach} explains the proposed approach in a general manner. Section \ref{subsec:Implementation} presents a proof of concept implementation for the proposed method, Section \ref{subsec:TestingScenario} presents the testing scenario and \ref{subsec:Results} the results. Implications of the results and current limitations are discussed in Section \ref{sec:Discussion}. In Section \ref{sec:FutureWork}, the authors explain how they plan to address these limitations in their future work, followed by concluding remarks in Section \ref{sec:Conclusion}
\begin{figure*}[h!]
	\centering
	\includegraphics[width=0.9\textwidth]{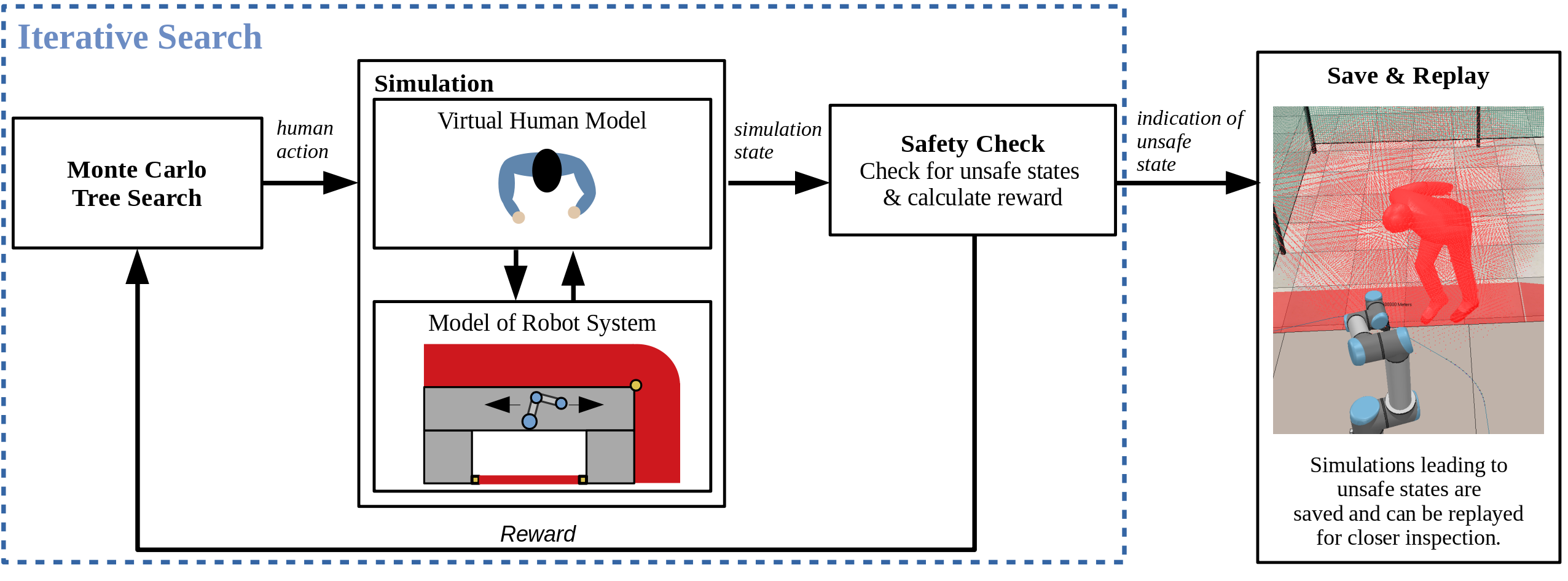}
	\caption{\small Iterative Search Procedure: The Monte Carlo Tree Search (MCTS) algorithm selects human actions which are carried out by a virtual human model in conjunction with a model of the robot cell. In each iteration, it is checked if the current simulation state is unsafe with respect to a user-defined set of safety-criteria, and a risk metric is calculated. By rewarding the occurence of unsafe states, this risk metric guides the MCTS algorithm towards creating dangerous situations in which unsafe states are likely to occur.}
	\label{fig:SystemOverview}
\end{figure*}
\section{RELATED WORK}
\label{sec:RelatedWork}
Safety engineering  typically relies on methods like “Hazard and Operability Analysis” (HAZOP) \cite{IEC61882}, "Failure Modes and Effects Analysis" (FMEA) \cite{IEC60812}, or “Systems-Theoretic Process Analysis” (STPA) \cite{Leveson2016} to identify hazards. These methods are semi-formal, that is, they define a certain hazard identification procedure but largely rely on human reasoning. They can be applied to a wide range of safety-critical systems.\newline
There are also several novel approaches that are specifically aimed at robotics: Guiochet proposed the use of HAZOP-UML, a HAZOP extension that uses UML diagrams, for analysis of robot systems \cite{Guiochet2016}. Marvel et al. have proposed task-based method that supports risk assessment using a ontology of HRC tasks \cite{Marvel2014}. Awad et al. have developed a rule-based expert system for risk assessment of HRC workplaces \cite{Awad2017}. Their tool allows the user to model the workplace using a model of products, processes, and resources ("PPR-model"). The PPR-model properties are mapped to hazards based on a set of pre-defined rules. The method "SAFER-HRC", developed by Askarpour et al. \cite{Askarpour2016,Askarpour2017,Askarpour2017a} and Vicentini et al. \cite{Vicentini2019}, uses formal verification methods for safety verification of HRC systems.\newline
While all of these methods are suitable to identify hazards in robot systems, they do have some limitations: Semi-formal methods rely largely on human reasoning and domain-specific knowledge and thus can be difficult to apply to novel and complex systems. Formal and rule-based approaches require a specific system model like the formal language description from \cite{Askarpour2016} or the PPR-model from \cite{Awad2017} which must be obtained specifically for the purpose of hazard identification. Furthermore, these models typically require significant modeling simplifications.\newline% i.e. there is a trade-off between precision and exhaustivenes (as emphasized for example in \cite{Vicentini2019}).\newline
An alternative approach that avoids these problems is simulation-based safety testing. In the field of robotics, simulation-based safety testing is typically done on a component level, e.g. for testing safety-critical control code. Examples for this are seen in the works of Araiza-Illan et al. \cite{Araiza2015,Araiza2016}, Bobka et al. \cite{Bobka2016}, and Uriagereka et al. \cite{Uriagereka2019}. In contrast, the use of simulation-based testing to identify hazards on a system level is still relatively unexplored. %Although Askarpour et al. recently presented a Co-simulation approach which couples SAFER-HRC with the simulator MORSE \cite{Askarpour2020}, they use MORSE primarily for visualization, while the hazard identification itself is conducted with an underlying formal model. %To the best of our knowledge, there are no purely simulation-based approaches that can  hazard identification in robot systems.
%who use model-based test generation for a human-robot handover task. Another promising simulation-based approach outside the domain of robotics is Adaptive Stress Testing (AST) \cite{Lee2018}, which uses Reinforcement Learning to find conditions where safety-critical systems are likely to fail. AST was applied successfully in the domains of aerospace systems by Lee et al. \cite{Lee2015} and autonomous driving by Koren et al. \cite{Koren2018}.

\section{PROPOSED APPROACH}\label{sec:Approach}
\subsection{Objective, Assumptions, and Basic Idea}
This paper explores a novel concept that uses simulation to find hazards in robot systems. As explained in the introduction, a major challenge is that in many cases hazards only manifest themselves in specific situations. In a dynamic simulation environment the number of possible simulation sequences can be vast. Thus, it can be difficul to create specifically those simulation sequences that lead to situations where existing hazards are uncovered.\newline
Our approach relies on the assumption that the behavior of the robot system is deterministic for a given human behavior. This means that if there are inherent hazards in the system, then there are certain human behaviors for which these hazards manifest themselves in form of an unsafe state, that is, an accident or near-accident. This assumption leads to the basic idea behind our approach: To expose hazards by creating high-risk human behavior that provokes accidents. To achieve this, we draw on the concept of AST \cite{Lee2018}: We use the Monte-Carlo Tree Search Algorithm (MCTS) from \cite{Lee2018} to control a virtual human model which is placed in a simulation model of the robot system under test. By optimizing the behavior of the virtual human to maximize a risk metric, the algorithm provokes unsafe situations. As our proof of concept will show, this approach can significantly increase the chance of finding hazards in simulation.\newline
\subsection{Problem Formulation}
Formally, the approach can be framed as a search problem where the goal is to find sequences of human actions that result in unsafe states. The search problem is described in a 5-tuple:
\begin{equation}
\langle S,U,A,\phi, s_0 \rangle
\end{equation} where $S$ is a set of simulation states that describe the combined configuration of the human model and the robot system model (including not only the robot itself but also other safety-related components, e.g. sensors). $U$ is a user-defined subset of $S$ that includes unsafe states, that is, states that violate a certain safety-condition formulated by the user. The set $A$ consists of the actions which can be performed by the human model in simulation. Note that in the following proof of concept example, $A$ is a set of simple human movement primitives. However, $A$ does not necessarily have to consist only of movements. It could also include other human actions that are relevant to the system under test, such as operator commands to the system. The function $\phi$ is a transition function that returns the next state given the current state and a human action: $s'=\phi(s,a)$. This function is implemented by the simulation, that is, the next state is obtained by simulating the interaction between human and robot system for a given human action.
Starting from the initial simulation state $s_0$, the goal is to find sequences of human actions $a_1, a_2, ... a_n$ which, when simulated in interaction with the robot system, result in an unsafe simulation state $s \in U$. The difficulty is that $U$ is only known implicitly: While it is easy for the user to define certain high-level safety constraints (e.g. \textit{"all collisions with the robot must be avoided"}), it is unknown what the specific system states are in which these constraints are violated and which action sequences lead to them.

\subsection{Search Procedure}
We solve this search problem with an iterative search procedure as shown in Fig. \ref{fig:SystemOverview}: The Monte-Carlo Tree Search (MCTS) algorithm iteratively selects a human action which is then carried out by the human model in interaction with the robot system model. After each action, the current simulation state $s$ is evaluated in a safety check to determine if an unsafe state $s\in U$ is reached. Furthermore, a reward $R$ is calculated and fed back to the MCTS algorithm. This reward is designed in a way that encourages dangerous behavior and thus accelerates the finding of hazards. If an unsafe state is reached, the simulation stops and the user can examine the hazard by replaying the simulation sequence that has led to the unsafe state. The user can then eliminate the hazard by implementing appropriate safety measures and restart the search with an updated simulation model to find further hazards. If desired, this process can be repeated throughout the whole system design stage.\newline
Note that the set $U$ of unsafe states depends on the safety condition that is defined by the user. Depending on the context of the application, one might define conditions based on criteria like velocity and distance (e.g. \textit{"all contact between human and robot must be avoided while the robot is moving with a velocity greater than X"}) or on collision characteristics like collision force and affected body part. (e.g. \textit{"all collisions that subject body part X to a collision force greater than Y must be avoided"}). For reasons of computational complexity, the following proof-of-concept example will use a simple velocity/distance criterion. In the future we will also include a collision force estimation into our method to allow for more sophisticated safety criteria.

\section{PROOF OF CONCEPT}
This section presents our proof of concept example: We use the MCTS algorithm of \cite{Lee2018} and the simulator CoppeliaSim (formerly known as {V-REP} \cite{Rohmer2013}) to implement the search procedure from Fig. \ref{fig:SystemOverview}. We then use this implementation to find hazards in an industrial robot cell. It should be noted that being a proof of concept, the presented implementation contains several simplifications which we will address in our future work.
\subsection{Implementation}
\label{subsec:Implementation}
%The implementation is based on the robotics simulator CoppeliaSim\footnote{(www.coppeliarobotics.com)}, formerly known as {V-REP} \cite{Rohmer2013}. It should be emphasized that the presented implementation is intended as a proof of concept and thus contains some simplifications, especially regarding the human model.\newline
\textbf{Human Model:} We use a simple human model from CoppeliaSim and augment it with additional joints so that it can perform a set of basic motions (five walking- and six upper-body motions, amounting to an action space of 30 combined motions):
\begin{dmath}\breakingcomma
	A_{Walking}=\left\{ (\mathrm{walk\ forward}),(\mathrm{turn\ left\ 45\degree}),(\mathrm{turn\ left\ 90\degree}),(\mathrm{turn\ right\ 45\degree}),(\mathrm{turn\ right\ 90\degree}) \right\}
	\label{eq:HumanActionStart}
\end{dmath}
\begin{dmath}\breakingcomma
	A_{UpperBody}=\left\{ (\mathrm{move\ body\ upright}),(\mathrm{bend\ forward)},(\mathrm{bend\ left)},\mathrm{(bend\ right)},\mathrm{(bend\ forward\ and\ right)},\mathrm{(bend\ forward\ and\ left)} \right\}
\end{dmath}
\begin{gather}
\label{eq:HumanActionEnd}
A=A_{Walking} \times A_{UpperBody}\\
|{A}|=5\cdot 6= 30
\end{gather}
Note that in our example, $A$ does not include arm motions. Arm motion is quite complex and representing it via explicit actions would likely lead to an explosion of the search space. Instead, we use an octree based on a reachable arm workspace computation \cite{Klopcar2005} to determine if the robot is within human reach. The parameters of the human model are shown in Table \ref{tab:ModelParameters}.
%\begin{figure}[t]
%	\centering
%	\includegraphics[width=0.7\columnwidth]{images/ArmModel.png}
%	\caption{Schematic depiction of the arm model used to compute the reachable arm workspace of the human model (adapted from \cite{Klopcar2005}): The model has five degrees of freedom: Flexion $\theta_{F}$ and Abduction $\theta_{A}$ of the shoulder, rotation $\theta_{R}$ of the upper arm, as well as flexion of the elbow ($\theta_E$) and the wrist ($\theta_W$). In the depicted home position, where all angles are zero, the upper arm is parallel to the $z-$axis, and the lower arm and hand are parallel to the $y-$axis.}
%	\label{fig:ArmModel}
%\end{figure}
\begin{table}[t]
	\begin{center}
		\begin{tabular}{|c||c||c|}
			\hline
			\textbf{Parameter} & \textbf{Value} & \textbf{Source}\\
			\hline
			Body Height & 1.78 m & Test person measurement \\
			\hline
			Upper arm length $l_U$& 0.30 m& Test person measurement\\
			\hline
			Lower arm length $l_L$& 0.31 m & Test person measurement\\
			\hline
			Hand length $l_H$& 0.18 m & Test person measurement\\
			\hline
			Walking speed & 1.6 $\mathrm{\frac{m}{s}}$ & Specified in \cite{ISO13855}\\
			\hline
			Max. Angle forward flexion & $\mathrm{55\degree}$& Derived from \cite{ISO13855}\\
			\hline
			Max. Angle lateral flexion & $\mathrm{35\degree}$& Specified in \cite{Medley2014}\\
			\hline
		\end{tabular}
	\end{center}
	\caption{Human Model Parameters}
	\label{tab:ModelParameters}
\end{table}

%\label{sec:Algorithm}
%\subsubsection{Algorithm}
\textbf{Algorithm:} To control the human model, we use the Monte-Carlo Tree Search algorithm\footnote{https://github.com/sisl/AdaptiveStressTesting.jl} from \cite{Lee2018}. For reasons of brevity, we only give a simplified explanation of the algorithm here. For a full explanation we refer to \cite{Lee2018}. The algorithm iteratively samples sequences of human actions from $A$ and executes them in the simulation. In keeping with the terminology of \cite{Lee2018}, we call these action sequences \textit{episodes}. After each action, it is checked if an unsafe state $s \in U$ has been reached. If this is the case, or if a maximum number of actions is reached, the episode terminates. The simulation is then set back to the initial state $s_0$ and a new episode begins. With each episode the algorithm incrementally expands a search tree, in which the edges correspond to human actions and the nodes to simulator states.\newline
We employ two variations of this algorithm: One basic version, which we call MCTS1, and one variation, which we call MCTS2. Whereas MCTS1 always starts its search at the initial simulation state $s_0$, MCTS2 commits to the most promising action after a certain number of episodes and uses the resulting simulation state as a new starting point. This results in a more exploitative search behavior.\newline
\textbf{Reward:} After each action, the algorithm receives a reward $R$. Based on the reward, a state-action value function is estimated which is used to adapt sampling of actions in future episodes. The reward should increase the chance of finding hazards by encouraging a more dangerous behavior of the virtual human. Thus, the occurrence of dangerous situations should be rewarded, whereas the occurrence of safe situations should be penalized. To quantify the level of danger that a situation holds, we define a safety index $c_S$:
\begin{equation}
c_S=(d_{HR}^2+1)\cdot e^{-v_R}
\end{equation}
where $d_{HR}$ is the human-robot distance and $v_R$ is the cartesian velocity of the fastest robot joint. The value of $c_S$ is large for \textit{safe} configurations (i.e. large distance, low speed). Since we want to encourage \textit{unsafe} situations, we give the inverse $\frac{1}{c_S}$ as a reward after each action. Additionally, we give the negative safety index $-c_S$ as penalty at the end of an episode when no unsafe state has been found. Thus, in total, the reward function is:
\begin{equation}
R=\left\{
\begin{array}{cc}
\frac{1}{c_S}, &\mathrm{if}\ k < n\\%\mathrm{\ and\ } s \notin U \\
-c_S, &\mathrm{if}\ k=n\mathrm{\ and\ } s \notin U
\end{array}
\right.
\end{equation}
where $k$ indicates the current step within the episode and $n$ is the episode length\footnote{Note that the reward structure differs from \cite{Lee2018}, where there is also a component that rewards the probability of actions. We changed this as we are interested in finding hazards independently of their probability.}.%Note that the search terminates when an unsafe state is found. Therefore, no reward is defined for $s \in U$.

\subsection{Test Scenarios}
\label{subsec:TestingScenario}
\begin{figure}
	\centering
	\includegraphics[width=0.8\columnwidth]{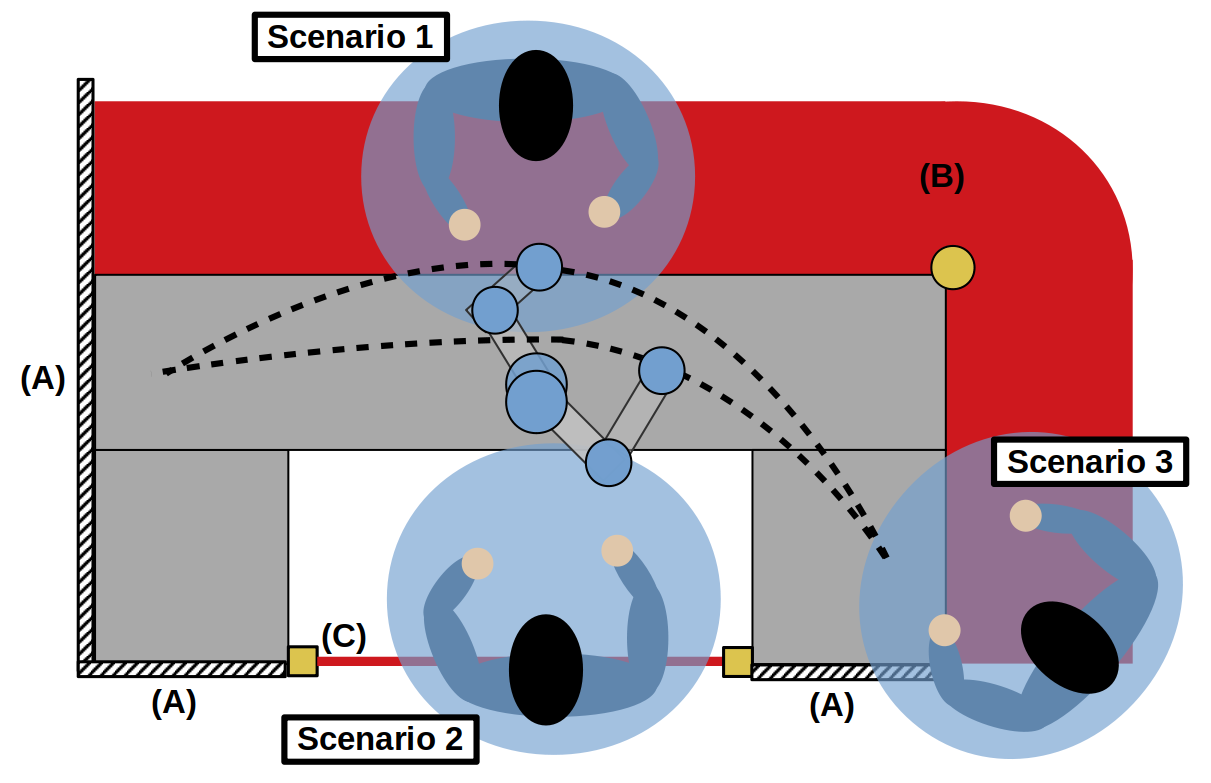}
	\caption{\small Top view of the robot cell, featuring safety fences (A), the laser scanner detection zone (B), and a light curtain (C). By introducing safety flaws into the cell design, three collision hazards (Scenario 1-3) were created.}
	\label{fig:HazardSituations}
\end{figure}

%\begin{figure*}
%	\centering
%	\begin{subfigure}[t]{0.43\textwidth}
%		\includegraphics[width=1\textwidth]{images/RobotCell.png}
%		\caption{View of the robot cell in CoppeliaSim}
%	\end{subfigure}
%	\hfill
%	\begin{subfigure}[t]{0.45\textwidth}
%		\includegraphics[width=1\textwidth]{images/HazardScenarios.png}
%		\caption{Top View of the robot cell with the collision hazards}
%	\end{subfigure}
%	\caption{\small Proof of Concept Test Scenarios.}
%	%    \label{fig:my_label}
%	\label{fig:HazardSituations}
%\end{figure*}

\begin{table*}[!h]
	\begin{tabularx}{\textwidth}{|X|p{8cm}|p{6cm}|}
		\hline
		\textbf{Test Scenario}& \textbf{Safety Flaw} & \textbf{Resulting Hazard}\\
		\hline
		\textbf{Scenario 1:}\newline Reduced width of\newline laser scanner zone & The width of the laser scanner protective field is reduced. Although the worker can still be detected by the laser scanner, the reduced field is too small to ensure that the robot stops completely before the worker can reach it. & A collision is possible if the worker approaches the table at the point where the robot path is closest and leans into the path as the robot passes (see \mbox{Fig. \ref{fig:HazardSituations}}, \mbox{Scenario 1}).\\
		\hline
		\textbf{Scenario 2:}\newline Altered robot path and position & Position and path of the robot are altered in such a way that the robot's elbow joint protrudes into the maintenance bay. Due to the protruding elbow joint, the distance between the light curtain and the robot is not sufficient anymore to stop the robot in time. & A collision is possible when the worker enters the maintenance bay (\mbox{Fig. \ref{fig:HazardSituations}, Scenario 2)}.\\
		\hline
		\textbf{Scenario 3:}\newline Partly removed safety fence & A part of the safety fence is removed. While the table itself is still closed off by the fence, the edge of the laser scanner field is not. & A collision can occur when the worker leans over the laser scanner field to reach around the remaining part of the safety fence (see \mbox{Fig. \ref{fig:HazardSituations}}, \mbox{Scenario 3)}.\\
		\hline
	\end{tabularx}
	\label{tab:Hazards}
	\caption{\small Description of Proof of Concept Test Scenarios}
\end{table*}
As a basis for the proof of concept tests we chose the industrial robot cell shown in Fig. \ref{fig:HazardSituations}. This cell combines typical safety features of industrial robot systems: Safety fences, a laser scanner, and a light curtain. In the center of the cell there is a U-shaped table on which the robot is mounted. The robot imitates a pick-and-place task between the two sides of the table. To intervene in the process, e.g. to refill parts, workers can approach the table either by walking through the laser scanner field, or by passing through a light curtain at the back of the cell. Areas not monitored via laser scanner or light curtain are closed off by the fences. Upon detection of a worker, laser scanner and light curtain send a stop signal to the robot. Note that due to the response time of the sensors, the stop signal is delayed. Furthermore, the robot needs a certain stopping time to reach a standstill. %A delay of \mbox{50 ms} for the light curtain and \mbox{350 ms} for the laser scanner is assumed. The stopping time of the robot is \mbox{150 ms}.
The cell is designed to satisfy the following safety condition: \textit{"Contact between human and robot must not be possible unless the robot stands still"}. Thus fences, laser scanner, and light curtain are configured in a way that even with the sensor delay and the robot stopping time, the worker cannot reach the robot before it has stopped \cite{ISO13855,ISO13857}. Since these safety measures should avoid any contact between human and robot while the robot is moving, the set of unsafe states $U$ in our example is defined as follows:
\begin{equation}
U=\{s\ |\ v_R>0, d_{HR}=0\}
\end{equation}
By altering the original cell layout and deliberately introducing safety-critical design flaws, we created three test scenarios where unsafe states are possible, each scenario containing a specific collision hazard. The scenarios are shown and explained in Fig. \ref{fig:HazardSituations} and Table II.\newline
Note that the movement of the robot, the sensor delays, and the robot stopping time make the scenario \textit{dynamic}. Although the dynamic effects here are relatively simple, they show that the method is able to find hazards in dynamic simulations and not only in static environments.

\subsection{Test Runs}
\label{subsec:Results}
\begin{figure*}[h!]
	\vspace{2mm}
	\includegraphics[width=1\textwidth]{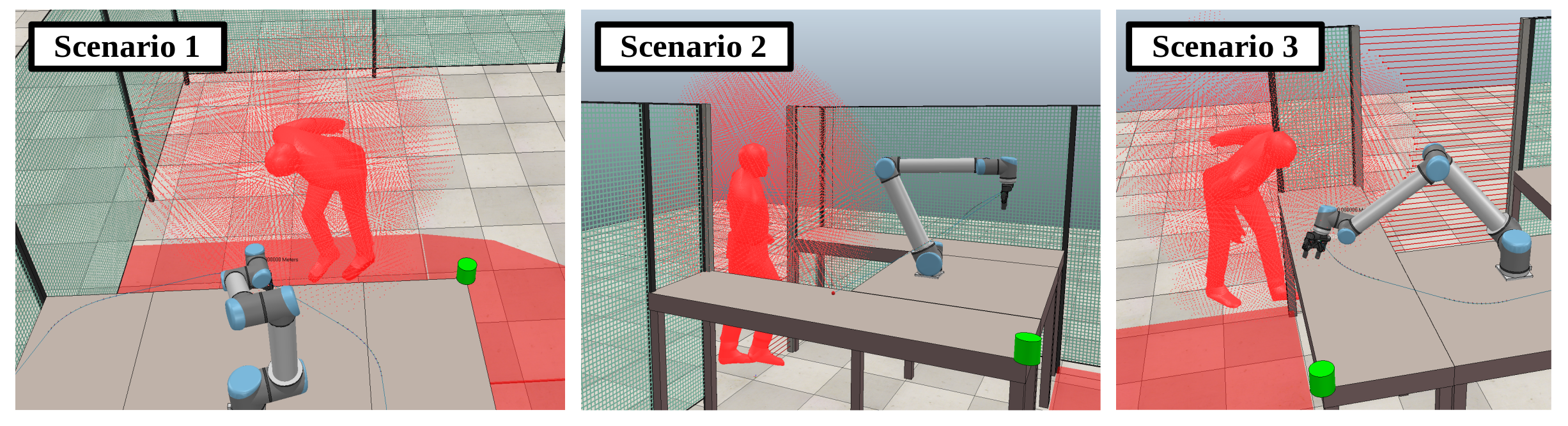}
	\caption{\small Hazard situations found in the three test scenarios (corresponding to Fig. \ref{fig:HazardSituations}). The red cloud indicates the volume reachable by the human. In all thee situations, the human is able to reach the robot while it is moving, and thus, the safety condition is violated.}
	\label{fig:HazardsIdentified}
\end{figure*}

\textbf{Setup: }Test runs are performed in CoppeliaSim with simulation timesteps of 50 ms. Each human action has a duration of four timesteps and each episode consists of eight actions. Test runs are conducted from two different starting points, one on the upper end of the cell for scenario 1 and one on the lower end of the cell for scenario 2 and 3 (compare Fig. \ref{fig:HazardSituations} for the test scenarios and Fig. \ref{fig:HazardsIdentified} for examples of corresponding hazard situations). Although this may seem like a convenient simplification, it is justifiable from a practical perspective, since a user would certainly select meaningful starting points and not place the human model at random. Each test scenario is performed with both MCTS1 and MCTS2. To show that our approach does indeed increase the chance of finding hazards, we conduct a random search for comparison in which episodes are assembled by sampling actions from a uniform distribution over $A$. For each combination of test scenario and algorithm, ten test runs are conducted with different random seeds. Each test run is limited to 200 episodes.

%\subsection{Results}
\textbf{Results: }Results are shown in Table III. The first row shows the success rates, i.e. in how many of the test runs the hazard was found. If no hazard is found within 200 episodes, a test run is considered unsuccessful. The second row shows the runtime, that is, the average number of episodes until discovery of the hazard (unsuccessful test runs are counted with the maximum of 200 episodes). It can be seen clearly that the two MCTS variants perform significantly better than the random search, both in terms of success rate and run time, which indicates that the adaptation of human behavior does indeed increase the chances the finding of hazards. However, it can also be seen that hazards can be missed. This is not only the case for the random search but also for both MCTS algorithms (although much less frequently). Meanwhile, comparing the MCTS variants with each other shows no clear advantage for either of them, especially given the small number of test scenarios. More tests will be conducted in the future to investigate potential differences in performance.
\section{DISCUSSION}
\label{sec:Discussion}
As the proof of concept has shown, the method can identify hazards in a realistic, industry-like robot system. Compared to a random search, it finds hazards significantly quicker and with a higher success rate. However, being in a proof of concept phase, there are several limitations to its applicability, especially the simplistic human model. Furthermore, in its current implementation, the method can only find one hazard at a time. In a practical application, the user would have to eliminate the found hazard by updating safety measures and then repeat the search to find further hazards. While this avoids the problem of local minima (i.e. discovering the same hazard repeatedly), it is impractical. Another, more fundamental limitation comes from the fact that the method is based on falsification of safety conditions. This means it cannot give a safety guarantee, it can only find counterexamples of situations where safety conditions are violated. Thus, it should be seen as an addition to existing methods rather than a replacement.\newline
The major advantage of the method is that it can find hazards autonomously while reducing the required amount of prior knowledge about the system to a minimum. Furthermore, it can be easily integrated into common robot simulator models which are widely used and does not require building a system model specifically for hazard analysis. These properties are highly desirable for the analysis of novel and complex systems. Since the proof of concept implementation is relatively simple, the full extent of these advantages may not be visible yet. However, we believe that there is great potential in this approach and that it could provide a powerful, scalable, and flexible tool for testing various types of complex robot systems, not only in the industrial context.

\begin{table}[t]
	\centering
	\begin{tabular}{|p{3cm}|c|c|c|c|}
		\hline
		& & \multicolumn{3}{|c|}{\textbf{Scenario}}\\
		\cline{3-5}
		& \textbf{Algorithm}& \textbf{1} & \textbf{2} & \textbf{3}\\
		\hline
		\hline
		\multirow{3}{*}{\textbf{Success rate}}
		& Random &  3/10  & 3/10 & 8/10 \\
		\cline{2-5}
		& MCTS1 & 10/10 & 8/10 & 10/10 \\ 
		\cline{2-5}
		& MCTS2 & 10/10 & 9/10 & 10/10 \\ 
		\cline{2-5}
		\hline
		\hline
		\multirow{3}{*}{\textbf{Avg. number of episodes}}
		& Random & 166.8 & 150.4 & 81.1\\
		\cline{2-5}
		& MCTS1 & 70.0 & 63.3 & 34.9 \\ 
		\cline{2-5}
		& MCTS2 & 49.7 & 80.4  & 38.0 \\ 
		\cline{2-5}
		\hline
	\end{tabular}
	\label{tab:Results}
	\caption{Results of the test runs} 
\end{table}
\section{FUTURE WORK}
\label{sec:FutureWork}
Currently, the method's limitations mainly result from simplifications in modeling and implementation. Especially the fact that we use a static octree for the arm workspace rather than an articulated arm model limits the types of hazards that can be identified. This will be addressed by augmenting the reachability model with an articulated arm model. Moreover, a collision force estimation will be incorporated. This will allow the method to test systems not only against velocity- and distance-based safety criteria, but also against collision force limits. Another aim is to enable a search for multiple hazards in one run. This will require adaptations to the MCTS to avoid convergence in local minima. To enable a widespread practical application, it should also be investigated how the method can be implemented in other common robot simulators, for example \textit{Visual Components}, \textit{ProcessSimulate}, etc. 

\section{CONCLUSION}
\label{sec:Conclusion}
A simulation-based method for safety testing of robot systems was proposed and evaluated. The method uses a human model and Monte Carlo Tree Search to find unsafe system states in simulation, which enables an automated hazard identification and reduces the reliance on prior system knowledge. A proof of concept has shown promising results, but the current implementation is still relatively simple and requires further development. Since the method is based on falsification of safety conditions, it cannot give a safety guarantee. Thus, it should be seen as an addition to existing methods rather than a replacement. %The capabilities of the current implementation are limited by some modeling simplifications, which will be addressed in the future.

\section*{ACKNOWLEDGMENT}
We thank the authors of \cite{Lee2018} for sharing AST on Github, and Gabriel Zerrer for his assistance with the simulation.


%%%%%%%%%%%%%%%%%%%%%%%%%%%%%%%%%%%%%%%%%%%%%%%%%%%%%%%%%%%%%%%%%%%%%%%%%%%%%%%%

\bibliography{IEEEabrv,preprint}
\bibliographystyle{IEEEtran}

\end{document}